\documentclass[journal,twoside,web]{ieeecolor}
\usepackage{tmi}
\usepackage{cite}
\usepackage{amsmath,amssymb,amsfonts}
\usepackage{algorithmic}
\usepackage{graphicx}
\usepackage{color}
\usepackage{textcomp}
\usepackage{array}
\usepackage{multirow}
\usepackage{booktabs}
\usepackage{indentfirst}

\usepackage{makecell}
\usepackage{diagbox}
\usepackage[misc]{ifsym}
\usepackage[colorlinks,linkcolor=blue]{hyperref}
\usepackage{caption}
\usepackage{dsfont}
\usepackage{makecell}
\usepackage{ulem}

\usepackage[ruled]{algorithm2e}

\newcommand{\wt}{\textcolor[RGB]{0,0,0}}
\newcommand{\major}{\textcolor[RGB]{0,0,0}}
\newcommand{\change}{\textcolor[RGB]{0,0,0}}

\def\BibTeX{{\rm B\kern-.05em{\sc i\kern-.025em b}\kern-.08em
		T\kern-.1667em\lower.7ex\hbox{E}\kern-.125emX}}
\begin{document}
	\title{STAR-RL: Spatial-temporal \\Hierarchical Reinforcement Learning for \\Interpretable Pathology Image Super-Resolution}
	\author{Wenting Chen, Jie Liu, Tommy W.S. Chow, \IEEEmembership{Fellow, IEEE}, Yixuan Yuan, \IEEEmembership{Member, IEEE}
		\thanks{This work was supported by Hong Kong Research Grants Council (RGC) General Research Fund 14204321, 14220622 and Hong Kong Innovation and Technology Commission Innovation and Technology Fund ITS/229/22 .\textit{(Corresponding author: Yixuan Yuan)}}
		\thanks{Wenting Chen and Jie Liu contributed equally to this work and are listed in alphabetical order.}
		\thanks{W. Chen, J. Liu and Tommy W.S. Chow are with the Department of Electrical Engineering,
			City University of Hong Kong, Hong Kong SAR, China. Y. Yuan is with the Department of Electronic Engineering,
			Chinese University of Hong Kong, and was with the Department of Electrical Engineering,
			City University of Hong Kong, Hong Kong SAR, China. (e-mail:
			{wentichen7-c, jliu.ee}@my.cityu.edu.hk, eetchow@cityu.edu.hk, yxyuan@ee.cuhk.edu.hk).}
	}
	\maketitle

	\begin{abstract}
		\change{Pathology image are essential for accurately interpreting lesion cells in cytopathology screening, but acquiring high-resolution digital slides requires specialized equipment and long scanning times.
			Though super-resolution (SR) techniques can alleviate this problem, existing deep learning models recover pathology image in a black-box manner, which can lead to untruthful biological details and misdiagnosis. Additionally, current methods allocate the same computational resources to recover each pixel of pathology image, leading to the sub-optimal recovery issue due to the large variation of pathology image.
			In this paper, we propose the first hierarchical reinforcement learning framework named Spatial-Temporal hierARchical Reinforcement Learning (STAR-RL), mainly for addressing the aforementioned issues in pathology image super-resolution problem.} We reformulate the SR problem as a Markov decision process of interpretable operations and adopt the {hierarchical} recovery mechanism in patch level, {to avoid sub-optimal recovery}. 
		Specifically, the higher-level spatial manager is proposed to pick out the most corrupted patch for the lower-level patch worker.
		Moreover, the higher-level temporal manager is advanced to evaluate the selected patch and determine whether the optimization should be stopped earlier, thereby avoiding the over-processed problem.
		Under the guidance of spatial-temporal managers, the lower-level patch worker processes the selected patch with pixel-wise interpretable actions at each time step.
		Experimental results on medical images degraded by different kernels show the effectiveness of STAR-RL. 
		Furthermore, STAR-RL validates the promotion in tumor diagnosis with a large margin and shows generalizability under various degradations. The source code is available at \url{https://github.com/CUHK-AIM-Group/STAR-RL}.
	\end{abstract}
	\begin{IEEEkeywords}
		Reinforcement Learning, Super-Resolution, Markov Decision Problem, Whole Slide Imaging.
	\end{IEEEkeywords}

	\begin{figure}[tp]
		\includegraphics[width=\linewidth]{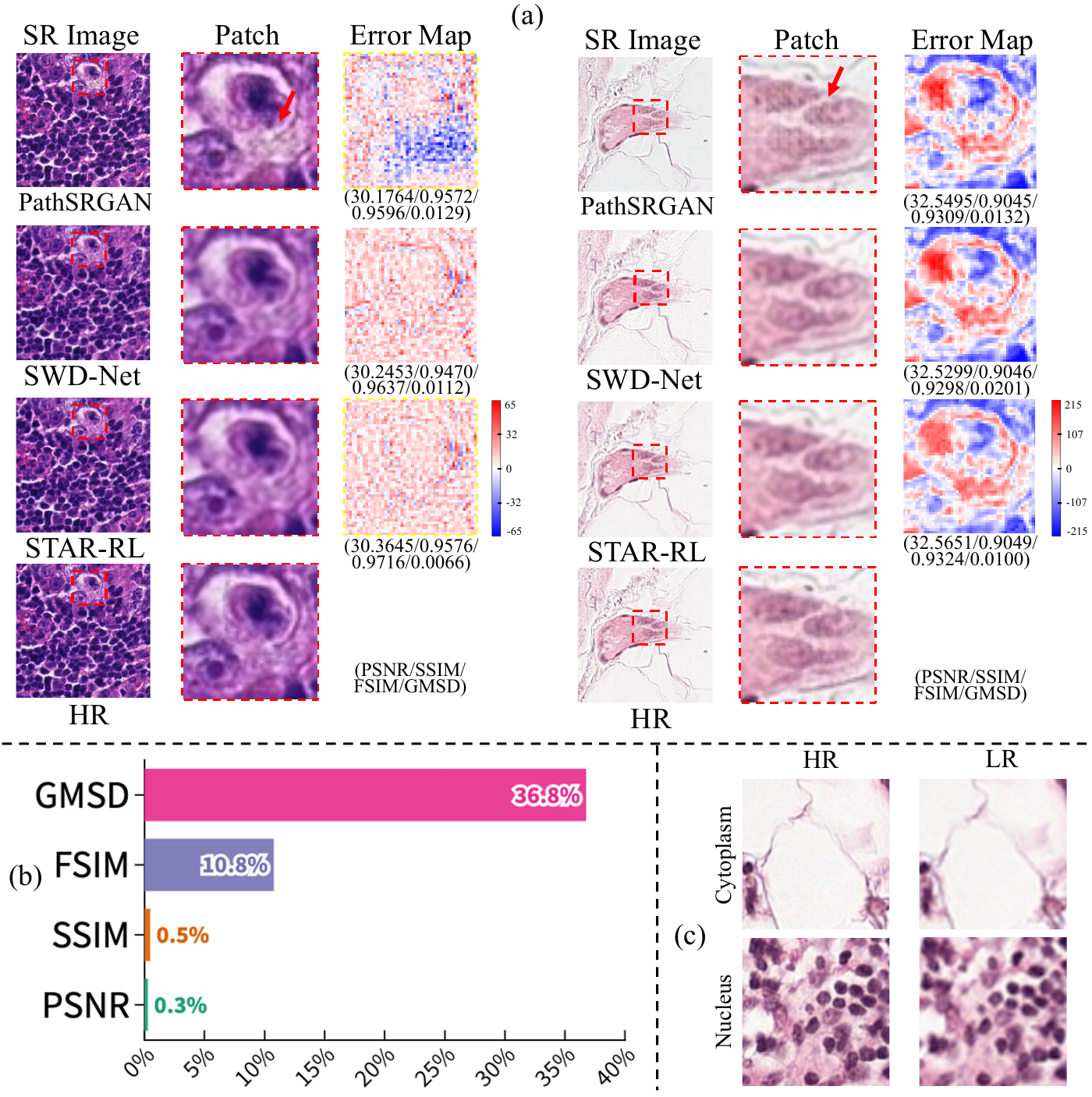} 
		\caption{\major{(a) The unexpected pattern generated by existing black-box method (PathSRGAN\cite{ma2020pathsrgan} and SWD-Net \cite{chen2020joint}); (b) SR performance gap with various metrics between black-box method and the proposed STAR-RL; (c) The cytoplasm and nucleus are representative cellular components. The cytoplasm with large-scale structure displays the smooth texture, which is easy to super-resolve. In contrast, the nucleus contains intricate biological details that are challenging to accurately restore due to significant information loss.}}
		
		\label{fig::bad_cases}
	\end{figure}
	
	\section{Introduction}
	\change{Pathology image is widely used to visualize the tissue sections for education, research, or diagnosis\cite{cornish2012whole,pantanowitz2011review}. In clinical application, the high-quality pathology scanners can produce pathology image for clinicians to locate and categorize abnormalities\cite{afshari2023single}. 
		However, the use of high-quality scanners can be complex, time-consuming, and costly due to specialized training, imposing additional financial burdens \cite{ghaznavi2013digital,park2019recent}. It would be beneficial to develop faster-speed and lower-cost devices for scanning tissues while maintaining image quality \cite{madabhushi2016image}. 
		One solution is to employ the deep learning technique \cite{srivastav2019human,liu2022edge} to convert low-resolution (LR) pathology image into high-resolution (HR) ones for better visual quality.}

	Automatic super-resolution (SR) algorithms have been investigated to super-resolve the LR images for several decades.
	Early methods, including neighbor embedding \cite{chang2004super,su2005neighborhood}, statistical image prior \cite{kim2010single,sun2008image} and sparse representation \cite{timofte2013anchored}, are based on carefully designed feature engineering and lack of flexibility. 
	Recently, deep learning-based {(DL)} super-resolution technology has rapidly developed from convolutional neural network (CNN) based methods \cite{kim2016accurate,tian2020coarse,liu2020residual} to generative adversarial network (GAN) based methods \cite{zhou2021ultrasound,liang2022details,liang2021hierarchical}. They are capable of capturing representative features automatically with data and mapping from LR to HR in the end-to-end fashion with satisfactory performance. 
	In the field of pathology image, some CNN-based \cite{mukherjee2019super,mukherjee2018convolutional,chen2020joint,chen2021super} and GAN-based \cite{ma2020pathsrgan,li2021single} methods are also proposed to improve the quality and clarity of biological structures via specially designed network structure, loss functions and other aspects. 
	
	Even though they have achieved great performance, there are still two pivotal challenges in current works for pathology image super-resolution. First, these works directly map the LR image to the HR one in black-box manner with uninterpretable mapping, leading to unforeseen patterns in the image details when processing new samples.
	As depicted in Fig. \ref{fig::bad_cases} (a), despite high structural similarity index (SSIM)\footnote{Structural similarity index (SSIM) globally assesses the image quality by comparing luminance, contrast and structure.}, {the SR image generated by PathSRGAN\cite{ma2020pathsrgan} shows artifacts, such as checkbox patterns, in the nucleus region. }
	Importantly, biological details such as heterochromatin and nuclear chromosomes are the basis for pathological experts to assess diseases \cite{riddle2011plasticity,donovan2021mitotic}. These inaccurately restored details would lead to incorrect observations of chromosomes and potentially result in misdiagnosis. 
	{Along this line, it is important to develop a whole new paradigm}
	to recover the resolution of pathology image with sequential classical operations, which are interpretable and avoid artifacts. In addition, better characterization metrics\footnote{Gradient Magnitude Similarity Deviation (GMSD) \cite{xue2013gradient} and feature similarity (FSIM) \cite{zhang2011FSIM} can measure the significance of local structure.} to reveal biological details should be employed to evaluate the pathology image quality.
	
	The second challenge lies in the sub-optimal detail recovery problem in the pathology image. \major{Existing pathology image super-resolution works treat the entire image uniformly, allocating the same amount of computation to each pixel to restore the SR image.} 
	In fact, the cytoplasm, which occupies most of the cell region, is smooth and main structures are preserved in the LR image, {as shown in Fig. \ref{fig::bad_cases} (c)}. On the other hand, the cell nucleus, which occupies less cell region, is non-flat and contains more biological details, such as nuclear shape and chromatin density \cite{kanavati2022deep,schapiro2022mcmicro}. Thus, there is a large variation of information degradation in pathology image, and non-flat regions with higher information density naturally lose more details. 
	{Allocating computational resources equally} would lead to sub-optimal recovery for pathology image. 
	This motivates us to design a framework to allocate more computational resources to non-flat regions and fewer resources for flat regions {to provide adequate effort to each region}.
	
	\begin{figure}[tp]
		\centering
		\includegraphics[width=0.95\linewidth]{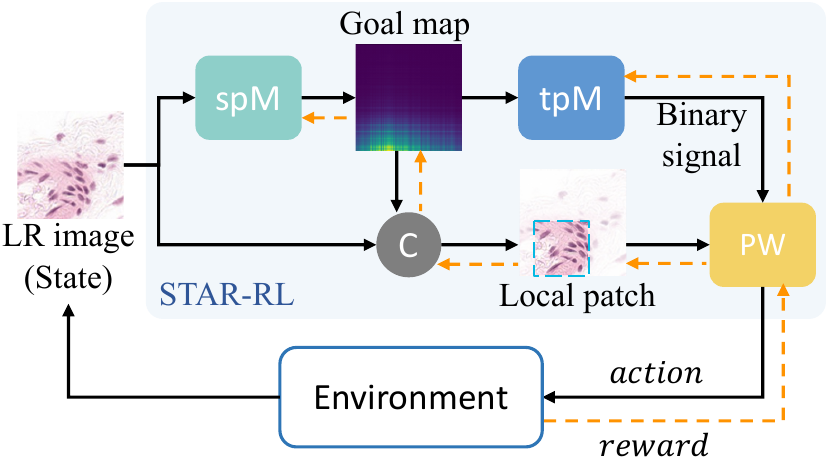}
		\caption{The overview of Spatial-Temporal hierARchical Reinforcement Learning (STAR-RL) framework. It includes higher-level spatial Manager (spM), temporal Manager (tpM) and lower-level patch worker (PW). The black solid and yellow dashed lines represent the forward and reward propagation. The grey circle with C indicates the patch crop operation.}\label{fig::overview}
	\end{figure}
	
	To address the abovementioned challenges, we present a framework named Spatial-Temporal hierARchical Reinforcement Learning (STAR-RL) for interpretable pathology image super-resolution, which reformulates image SR problem as the Markov decision process and attempts to tackle it with hierarchical reinforcement learning. As shown in Fig. \ref{fig::overview}, we reformulate the SR problem as a Markov decision process. Given the current state (i.e. LR image), STAR-RL decides the action (i.e. classical understandable operations \cite{gonzalez2009digital}) to recover the image and obtain reward from environment (i.e. error measurement) to learn sequential decision process. This reformulation makes the whole SR process interpretable and transparent. Based on this reformulation, we propose a novel two-level hierarchical mechanism to restore the LR image in patch level instead of whole image. More specifically, (1) the \textbf{higher-level spatial Manager (spM)} takes the LR image as input to predict the goal map, which picks the worst corrupted patch for lower-level patch worker. (2) The \textbf{lower-level Patch Worker (PW)} selects the pixel-wise actions for local patch at every time step by following the goal from the spM. (3) The \textbf{higher-level temporal Manager (tpM)} determines whether the goal is completed and when to terminate the optimization. In other words, the spM decides which patch should be recovered, and the PW decides the actions to restore the patch sequentially. The tpM is employed to evaluate whether the goal is completed. This hierarchical mechanism establishes a {reasonable and effective} resource allocation mechanism {to avoid sub-optimal recovery}. To be summarized, the main contributions of our work are as follows:
	
	\begin{figure*}[tp]
		\centering
		\includegraphics[width=0.9\textwidth]{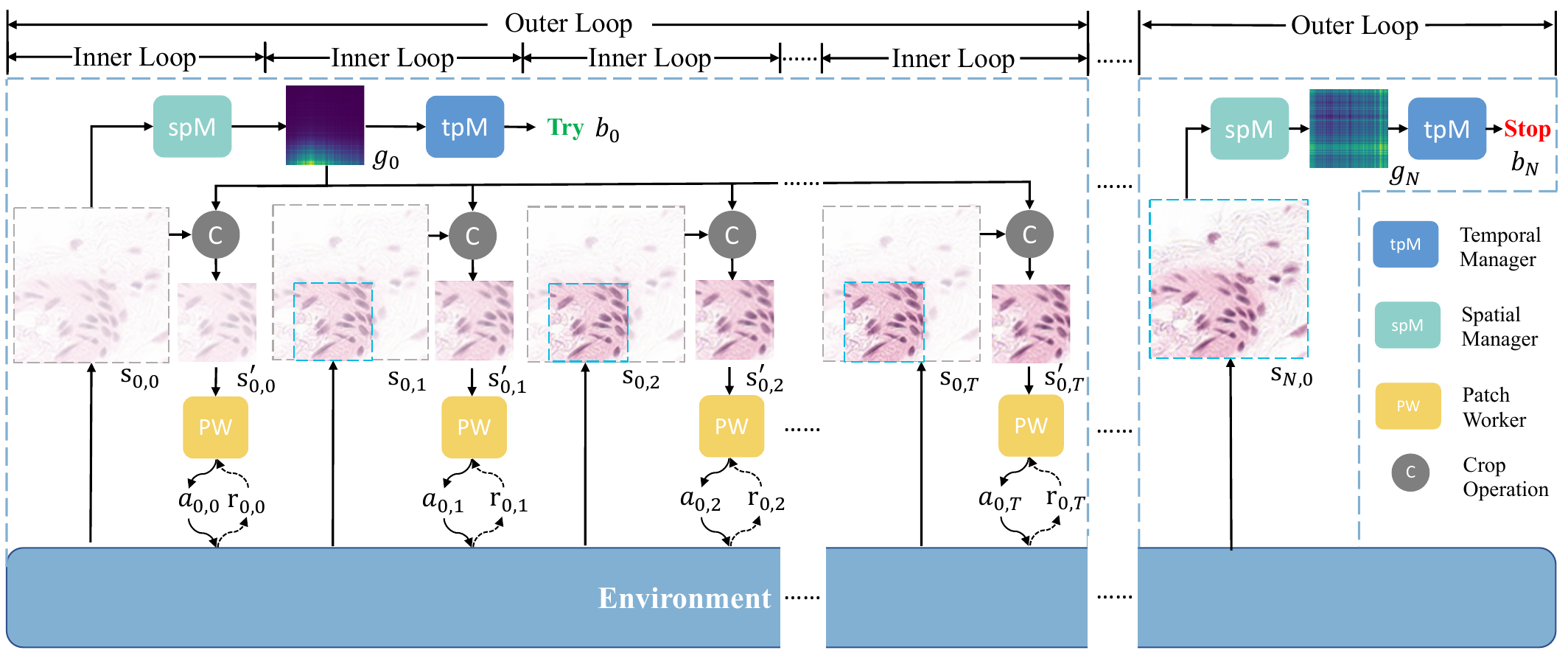}
		\caption{The overview of unrolled Spatial-Temporal hierARchical Reinforcement Learning (STAR-RL) framework, consisting of an outer loop for patch selection and several inner loops for patch recovering.}  \label{fig::framework}
	\end{figure*}
	
	\begin{itemize}
		\item We propose a Spatial-Temporal hierARchical Reinforcement Learning (STAR-RL) framework, which solves pathology image SR with sequential operation based on hierarchical reinforcement learning. To the best of our knowledge, this work is the first attempt to utilize the Markov decision process for pathology image super-resolution. 
		\item For interpretability, we advance a new paradigm to restore pathology image with sequential classical operations, which are interpretable and avoid artifacts.
		\item {For the sub-optimal detail recovery issue}, we propose a novel two-level hierarchical mechanism to restore the LR image in patch level with spM, tpM and PW. 
		\item Beyond previous commonly used global metrics PSNR and SSIM, we adopt the GMSD and FSIM to assess the quality of local biological details. Qualitative and quantitative results prove the superiority of STAR-RL framework in pathology image SR. In addition, ablation study is conducted to validate the effectiveness of each designed component. Furthermore, we validate the interpretability and generalizability of STAR-RL framework. At last, we evaluate the promotion of our method in tumor diagnosis tasks.
	\end{itemize}
	
	\section{Related Work}
	\subsection{SR for Medical Images.}
	\major{With the significant progress of deep learning (DL) in computer vision tasks, various medical image SR methods have begun to adopt deep neural network for super-resolution}, such as the convolution-based \cite{kim2016accurate,chen2021super,chen2020joint,chen2022dynamic,mukherjee2019super,chen2024unsupervised,zhao2019channel,li2019two} and GAN-based \cite{zhou2021ultrasound,almalioglu2020endol2h,upadhyay2019mixed,chen2022dynamic,chen2024mask} methods. For instance, EndoL2H \cite{almalioglu2020endol2h}, a GAN (Generative Adversarial Nets) \cite{goodfellow2014generative} based framework, introduced an attention U-Net to map the LR to HR image and a discriminator to distinguish the SR from HR image. 
	Ma et al. \cite{ma2020pathsrgan} proposed a GAN-based progressive multi-supervised SR model, called PathSRGAN, which designed a two-stage generator to learn the mapping of LR to HR cytopathological images and a discriminator to differentiate the generated images.
	SWD-Net proposed by Chen \textit{et al.} \cite{chen2020joint} employed a two-stage framework, including a spatial stage to iteratively rectify the image features and a wavelet stage to enhance the structural boundaries of SR images.
	MRC-Net introduced by Chen \textit{et al.} \cite{chen2021super} includes a MRC module for extracting both global and local features, as well as a refined context fusion module for effectively interacting with contextual information.
	Differently, the proposed STAR-RL aims to solve the super-resolution problem with interpretable Markov decision problem rather than black-box mapping manner.
	
	\subsection{RL for Image Processing.}
	Several pioneering works \cite{yu2018crafting,zhang2018dynamically,furuta2019fully,li2020mri} have adopted reinforcement learning for image processing tasks, such as image restoration, denoising, and color enhancement.
	\major{For example, Yu \textit{et al.} \cite{yu2018crafting} used deep reinforcement learning to select appropriate small-scale convolutional networks of varying complexities for image restoration.} 
	Zhang \textit{et al.} \cite{zhang2018dynamically} employed RL to dynamically control the decision loop time and recover images with various types of noise.
	\major{Hui \textit{et al.} \cite{hui2021learning}} utilized deep reinforcement learning to solve the non-differentiable optimization problem for blind SR, which adopted an actor to predict the blur kernel and an AMNet to fuse the blur kernel with LR image features for multiple degradations SR. Vassilo \textit{et al.} \cite{vassilo2020multi} introduced a multi-agent reinforcement learning method for SR, which adopts multiple agents to choose the predefined actions, such as the GAN-based SR algorithms to update pixel values. 
	Furuta \textit{et al.} \cite{furuta2019fully} and Li \textit{et al.} \cite{li2020mri} have employed reinforcement learning with a Markov decision process model for various image processing tasks, including denoising, restoration, and local color enhancement. \major{These methods have also applied multi-agent techniques by treating each pixel as an agent. To build upon these previous works, we introduce the concepts of spatial and temporal managers and patch workers to the existing reinforcement learning framework. These extensions enable efficient processing of image patches and facilitate the super-resolution of whole slide images. Additionally, our method applies basic and interpretable filters at the pixel level of the hierarchical framework, which enhances the interpretability of our approach.}

	\section{Method}
	\subsection{Problem Definition} 
	In this paper, our pathology image SR problem follows the definition of a Markov decision process (MDP), which includes state space $\mathcal{S}$, goal space $\mathcal{G}$, action space $\mathcal{A}$, reward function $\mathcal{R}$, and discount factor $\gamma \in [0,1)$. In our definition, LR image is viewed as state $s \in \mathcal{S}$, based on which higher-level and lower-level agents decide the goal $g \in \mathcal{G}$ and action $a \in \mathcal{A}$ in collaboration to recover the HR image. \major{Then, the environment measures the error with the HR image and returns feedback in the form of the reward $r \in \mathcal{R}$ to agents.}
	The objective of MDP is to maximize the mean of total expected rewards with a discount factor, i.e., to decide the optimal sequential actions. 
	
	\begin{table}[t]
		\centering
		\caption{\major{Summary of notation used in this work.}}
		\renewcommand{\arraystretch}{1.4}
		\begin{tabular}{ccc}
			\toprule[1pt]
			Index & Notation & Description \\ \hline
			1 & $s_{n, 0}$  & \makecell[l]{The SR image in the $n^{th}$ outer loop.}   \\ \hline
			2 & $s_{n, t}^{'}$  & \makecell[l]{The patch selected by spM in the $n^{th}$ outer \\loop and $t^{th}$ inner loop.}   \\ \hline
			3 & $g_{n}$  & \makecell[l]{The goal map for the $n^{th}$ outer loop.}   \\ \hline
			4 & $a_{n, t}$  & \makecell[l]{The action predicted by PW in the $n^{th}$ outer\\ loop and $t^{th}$ inner loop.}   \\ \hline
			5 & $r_{n, t}$  & \makecell[l]{The reward gained from environment in the\\ $n^{th}$ outer loop and $t^{th}$ inner loop.}   \\ \hline
			6 & $b_{n}$  & \makecell[l]{The binary signal for the $n^{th}$ outer loop.}   \\ \hline
			\major{7} & \major{$p$}  &\major{ \makecell[l]{The learnable parameters for filters.} }  \\ \hline
			\major{8} & \major{$y$} &\major{ \makecell[l]{The high resolution (HR) image.} }  \\ \hline
			\major{9} & \major{$y'$} &\major{ \makecell[l]{The high resolution (HR) image patch.} }  \\ \hline
			
			\toprule[1pt]
		\end{tabular}
		\label{tab::notation}
	\end{table}
	
	\begin{table}[t]
		\centering
		\caption{Predefined action list, i.e., action space $\mathcal{A}$.}
		\scalebox{0.9}{
			\renewcommand{\arraystretch}{1.3}
			\begin{tabular}{p{0.45cm}<{\centering}| p{2.2cm}<{\centering}| p{0.6cm}<{\centering}| p{1.3cm}<{\centering}| p{1.7cm}<{\centering}}
				\toprule[1pt]
				Index & Action              & Filter size & Parameter & Effect \\ \hline
				1& Sobel filter (left)  &   $3\times3$   & learnable & \multirow{4}{*}{\makecell[c]{Edge\\ enhancement}} \\\cline{1-4}
				2& Sobel filter (right) &   $3\times3$ & learnable \\\cline{1-4}
				3& Sobel filter (up)    & $3\times3$ & learnable \\\cline{1-4}
				4& Sobel filter (up)    &  $3\times3$    & learnable \\\hline
				5& \makecell[c]{Laplace filter}       &   $3\times3$    & learnable & Contrast enhancement \\\hline
				6& Gaussian filter      &   $5\times5$ & $\sigma=0.5$ & Smoothing\\ \hline
				7& Sharp              & $5\times5$  & $\sigma=0.5$, learnable & Sharpening \\ \hline 
				8& Addition             & $1\times1$ & 1 & \multirow{2}{*}{\makecell[c]{Fine-grained\\ modification}}\\\cline{1-4}
				9& Subtraction          & $1\times1$ & 1 \\\hline
				10& \makecell[c]{Do nothing}    & - & -  & Avoid over-process        \\\hline
				\toprule[1pt]
			\end{tabular}
		}
		\label{tab::action}
	\end{table}
	
	\subsection{Overview} 
	To make the illustration more clear, we summarize the important notations in Table \ref{tab::notation}.
	As shown in Fig. \ref{fig::framework}, the proposed Spatial-Temporal hierARchical Reinforcement Learning (STAR-RL) framework consists of multiple outer loops for patch selection and inner loops for patch recovery, where the higher-level spatial manager (spM), temporal manager (tpM) and lower-level patch worker (PW) collaborate hierarchically. 
	In an outer loop, given an \major{SR image $s_{n,0}~(n \in \{0,\dots,N\})$}, the spM predicts the goal map $g_n$ to select the most corrupted patch for the PW to restore. \major{The $N$ represents the maximum episode for outer loop.}
	Then, the tpM evaluates the goal map to determine whether the outer loop should be stopped and send a binary signal $b_n$ for early stop.
	In an inner loop, we perform crop operation on the LR image with the goal map and obtain the cropped patch $s_{n,t}^{'}$.
	With the supervision of two managers, 
	the PW is fed with $s_{n,t}^{'}$ and estimates the pixel-level actions $a_{n,t}~(t \in \{0,\dots,T\})$ to restore the selected patch. \major{$T$ represents the maximum episode for the inner loop.}
	\major{Finally, after interacting with environment, the reward $r_{n,t}$ calculated using HR image supervises the learning of the outer and inner loops.}

	\subsection{Hierarchical Mechanism}
	\subsubsection{The Outer Loop} 
	\major{The higher-level spM receives a whole image $s_{n,0} \in \mathcal{S}$ and utilizes the higher-level policy $\pi(g|s)$ to yield a 2D goal map $g_{n} = g_{n}^{h} \otimes g_{n}^{v}, \{g_n^h, g_n^v\} \in \mathcal{G}, n \in \{0,\dots,N\}$, where $n$ represents the time step of outer loop and $N$ indicates the maximum iteration number. The 2D goal map $g_n$,  represented by the outer product of two 1D horizontal and vertical vectors $g_n^h, g_n^v$,  shows the position of the most corrupted patch in state $s_{n,0}$.}
	Then the higher-level tpM determines whether the goal is completed and outputs a binary signal $b_n$ for early stop. 
	\major{The spM aims to maximize the expected total rewards  of outer loop $R_n$.
		\begin{align}
			R_n = \sum_{i=n}^{N}\gamma^i r_i, ~ r_i=|s_{i,0}-y|-|s_{(i+1),0}-y|,
		\end{align}
		where $r_i$ measures the decreased error compared with the previous one according to HR image $y$.}
	\major{As for tpM, the ground truth of binary signal is $0$ when the accumulated reward of outer loop $r_i$ is negative, indicating the optimization should be stopped, otherwise the signal should be $1$.}

	\subsubsection{The Inner Loop.} 
	\major{Formally, according to the given goal map $\{g_n^h, g_n^v\}$, the local patch $s_{n,t}'$ is cropped at location $(\arg\max_a g_n^h(a), \arg\max_b g_n^v(b) )$ with predefined patch size, where $t$ indicates the time step of inner loop, $a$ and $b$ are the value index of $g_n^h$ and $g_n^v$, respectively. $h$ and $v$ are the abbreviation of \textbf{h}orizontal and \textbf{v}ertical, respectively. }The lower-level PW receives the state $s_{n,t}'$ and applies a discrete policy $\pi(a|s)$ to decide the optimal action $a_{n,t} \in \mathcal{A}$ for each pixel.
	The action space $\mathcal{A}$ consists of classical interpretable filters in digital image processing \cite{gonzalez2009digital}, as listed in Table \ref{tab::action}. The Sobel filters in four directions (up, down, right, left) and sharp filter are typical sharpening filters for details enhancement, amplifying the contrast between the pixels. The Laplace filter is typical for contrast enhancement. 
	To avoid the over-processed content, Gaussian filter with the size of $5 \times 5$ is employed to smooth over-sharpened pixels.  
	\major{Also, the lower-level PW applies a continuous policy $\pi(p|s)$ to estimate the parameters $p$.}
	\major{The expected total rewards of inner loop $R_{n,t}$ are 
		\begin{align}
			R_{n,t} = \sum_{i=t}^{T} \gamma^i r_{n,i}, ~ r_{n,i} = |s_{n,i}' - y'| - |s_{n,(i+1)}' - y'|,
		\end{align}
		where $r_{n,i}$ measures the decreased error compared with the previous one according to HR patch image $y'$.}

	\begin{figure}[t]
		\centering
		\includegraphics[width=0.70\linewidth]{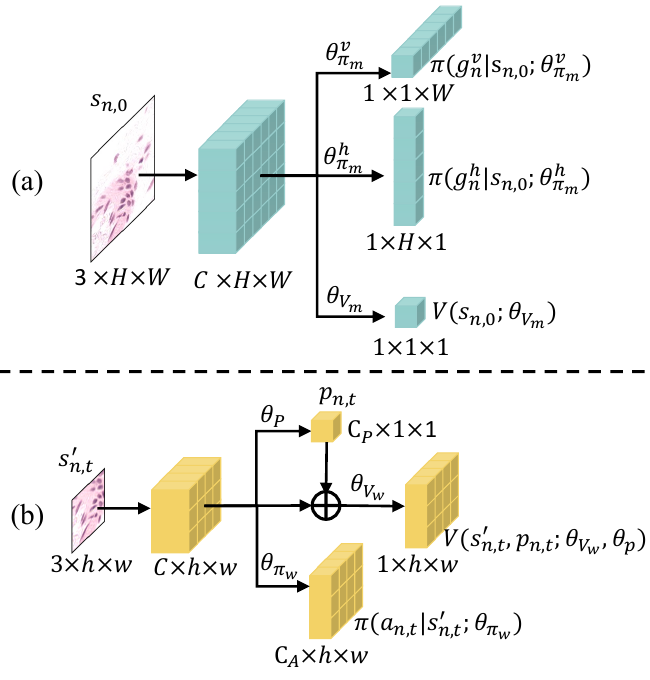}
		\caption{The architecture of (a) higher-level spatial manager and (b) lower-level patch worker. } \label{fig::network}
	\end{figure}
	

	\subsection{Network Architecture} 
	
	
	\subsubsection{The higher-level spM} As shown in Fig. \ref{fig::network} (a), in the higher-level spM, we first extract features of the LR image $s_{n,0}$ with two convolutional layers. To preserve the spatial information in each axis and predict the coordinate, we perform the axis-aware average pooling, i.e. horizontal pooling with window $(W, 1)$ and vertical pooling with window $(1, H)$ followed by two policy networks to determine the horizontal goal policy $\pi(g_n^h|s_{n,0};\theta^h_{\pi_m})$ and vertical goal policy $\pi(g_n^v|s_{n,0};\theta^v_{\pi_m})$, respectively. \major{The $W$ and $H$ means the width and height of image. The $\theta^h_{\pi_m}$ and $\theta^v_{\pi_m}$ represents the parameters of two policy networks for manager.} Each policy network includes a $1 \times 1$ convolutional layer. \major{Afterward, the value network outputs the value of state $V(s_{n,0};\theta_{V_m})$, which refers to the expected sum of future rewards that can be obtained from that state following a particular policy. The $\theta_{V_m}$ represents the parameters of the value network for the manager.}
	
	\subsubsection{The higher-level tpM} 
	The higher-level tpM utilizes the \major{multilayer perceptron (MLP)} structure followed by a Sigmoid function to take the goals $\{g_n^h, g_n^v\}$ as input to predict the binary signal $b_n$, which is described as $b_n=sigmoid(MLP([g_n^h, g_n^v]))$.
	
	\SetKwInOut{KwIn}{Require}
	\SetKw{training}{Training Algorithm}
	\SetKw{inference}{Inference Algorithm}
	\SetKw{Continue}{Finish Recovery}
	\begin{algorithm}[t]
		\LinesNumbered
		\caption{STAR-RL Training Algorithm} \label{algorithm} 
		\KwIn{Training pairs $\langle$LR images $s$, HR images $y$$\rangle$}
		Randomly initialize the model parameters $\theta$\;
		\For{each training pair $\langle$LR images $s$, HR images $y$$\rangle$}{   
			Set LR image $s$ as initial state $s_{0,0}$\;
			\For{n = 0 to N-1}{ 
				\tcp{Outer Loop}
				Feed spM with state $s_{n,0}$\;
				Obtain the goal map $g_n$\;
				Feed tpM with $g_n$ to get binary signal $b_n$\;
				\For{ t = 0 to T-1}{ 
					\tcp{Inner Loop}
					Crop patches $s_{n,t}^{'}$ based on $g_n$ from $s_{n,t}$\;
					Feed PW with $s_{n,t}^{'}$ to get action $a_{n,t}$\;
					Environment update state to $s_{n,t+1}$ and calculate reward for single inner loop $r_{n,t}$\;
				}
				Compute total reward of inner loop $R_{n,t}$\;
				Optimize PW with $L_{P_{\mathcal{W}}}$, $L_{V_{\mathcal{W}}}$ and $L_{\pi_{\mathcal{W}}}$\;
				Environment computes reward of outer loop $r_n$\;
			}
			Compute total reward for outer loop $R_n$\;
			Optimize the policy for spM with $L_{V_{m}}$ and $L_{\pi_{m}^d}$\;
			Update tpM with $L_{tem}$\;
		}
	\end{algorithm}
	
	\begin{algorithm}[h]
		\LinesNumbered
		\caption{STAR-RL Inference Algorithm} \label{algorithm2} 
		Set LR image $s$ as initial state $s_{0,0}$\;
		\For{n = 0 to N-1}{ 
			Feed spM with state $s_{n,0}$\;
			Obtain the goal map $g_n$\;
			Feed tpM with $g_n$ to get binary signal $b_n$\;
			\If{ $b_n < 0$ } { 
				\Continue 
			}
			\For{ t = 0 to T-1}{ 
				Crop patches $s_{n,t}^{'}$ based on $g_n$ from $s_{n,t}$\;
				Feed PW with $s_{n,t}^{'}$ to get action $a_{n,t}$\;
				Environment updates state to $s_{n,t+1}$\;
			}
		}
	\end{algorithm}
	
	\subsubsection{The lower-level PW} 
	As depicted in Fig. \ref{fig::network} (b), the lower-level PW extracts the features of the selected patch through four stacked convolutional layers. Then, the extracted features are fed to discrete policy network $\theta_{\pi_\mathcal{W}}$ and continuous policy network $\theta_p$ to learn the discrete action $a_{n,t}$ and the continuous parameter $p_{n,t}$ for parameterized operation. \major{Both $\theta_{\pi_\mathcal{W}}$ and $\theta_p$ employ convolutional neural networks.} The output of continuous action branch $p_{n,t}$ is set as the parameter of corresponding operations. The number of parameterized operation is defined as $C_{P}$.
	The $p_{n,t}$ is also fed into the value network $\theta_{V_\mathcal{W}}$ with $s'_{n,t}$ to evaluate the performance of continuous policy network.
	
	\subsection{Policy Learning} 
	In this work, we learn the policy with actor-critic algorithm \cite{sutton2018reinforcement}.
	This algorithm jointly trains the higher-level policy function $\pi(g|s)$ and the value function $V(s)$. 
	The loss functions for higher-level spM are shown below:
	\begin{gather}
		L_{V_{m}} = ||R_n - V ( s_{n, 0}; \theta_{V_m})||^2, \\
		L_{\pi_{m}^d} = - (R_n - V(s_{n,0}; \theta_{V_m})) \log \pi(g_n^d|s_{n,0};\theta_{\pi_m}^d), d \in \{h, v\},
	\end{gather}
	where $\theta_{V_m}$ and $\theta_{\pi_m}^d$ represent the parameters of value network and policy network for higher-level spM.
	The first term, corresponding to the value network, minimizes the temporal-difference error, while the second term rectifies the predicted probability vector from the policy network. With the supervision of these two loss functions, the higher-level spM can select the most corrupted patch of whole image that can bring larger reward.

	To optimize the higher-level tpM, we propose to maximize the log likelihood of $b_n$ according to inner loop reward summation with binary cross entropy:
	\begin{equation}
		L_{tem} = \mathds{1}_{[\sum_{t=1}^{T}r_{n,t}>0]} \cdot \log b_n + (1-\mathds{1}_{[\sum_{t=1}^{T}r_{n,t}>0]}) \cdot \log (1-b_n),
		\label{eq:tem}
	\end{equation}
	where $\mathds{1}_{[\cdot]}$ represents the indicator function. 
	Based on this, the higher-level tpM can learn when to stop the recovery process for preserving recovered details.
	
	Due to the continuous policy learning in lower-level PW, we optimize the parameter policy $\pi(p|s)$, 
	\begin{equation}
		L_{P_\mathcal{W}} = - V(s_{n,t}', p_{n,t}; \theta_{V_\mathcal{W}}, \theta_P).
	\end{equation}
	This loss aims to maximize the value of continuous action through only updating the continuous policy network parameter $\theta_P$. The value function $V(s)$ and lower-level discrete policy function $\pi(a|s)$ in lower-level PW is optimized with $L_{V_\mathcal{W}} = ||R_{n,t} - V ( s_{n, t}'; \theta_{V_\mathcal{W}})||^2$ and $L_{\pi_\mathcal{W}} = - (R_{n,t} - V(s_{n,t}'; \theta_{V_\mathcal{W}})) \log \pi(a_{n,t}|s_{n,t}';\theta_{\pi_\mathcal{W}})$, which is similar to Equation (1) and (2).
	With predefined actions and the objective function for each module, the lower-level PW can precisely restore the given LR image patches with interpretable classical image filters.
	
	
	\begin{table*}[th]
		\renewcommand{\arraystretch}{1.3}   
		\setlength\tabcolsep{4.5pt}  
		\centering
		\caption{\major{SR performance comparison with different methods on HistoSR dataset with BI degradation and GB degradation. The best and second best results are \textbf{highlighted} and \underline{underlined}.}}
		\begin{tabular}{c|c|c|c|c|c|c|c|c|c|c|c|c|c}
			\toprule[1pt]
			\multirow{2}{*}{Methods}& \multirow{2}{*}{Type}&\multicolumn{4}{c|}{BI Degrad.}   &\multicolumn{4}{c|}{GB Degrad.}  &\multirow{4}{*}{\makecell[c]{Params\\ (MB)}}&\multirow{4}{*}{\makecell[c]{FLOPs\\ (GB)}}&\multirow{4}{*}{\makecell[c]{Memory\\ (MB)}}&\multirow{4}{*}{\makecell[c]{\wt{Time}\\ \wt{(s)}}}\\ \cline{3-10}
			
			& & \multicolumn{2}{c|}{Global metrics}& \multicolumn{2}{c|}{Local metrics} & \multicolumn{2}{c|}{Global metrics}& \multicolumn{2}{c|}{Local metrics} & & & & \\ \cline{3-10}
			&  & PSNR $\uparrow$ & SSIM $\uparrow$ & FSIM $\uparrow$  & GMSD $\downarrow$ & PSNR $\uparrow$ & SSIM $\uparrow$ & FSIM $\uparrow$  & GMSD $\downarrow$& & & & \\\hline
			Bicubic & - & 28.847  & 0.897 & 0.9225  & 0.0158 & 28.179 & 0.881 & 0.9101  & 0.0229 & - & - & - & - \\\hline
			VDSR \cite{kim2016accurate}& & 31.362 & 0.895& \underline{0.9574}  & \textbf{0.0012} & 30.925 & 0.895 & \underline{0.9537}  & \underline{0.0015} & 0.67  & 24.68  & 343.27 & \textbf{\wt{0.020}}  \\\cline{1-1} \cline{3-14}
			PathSRGAN\cite{ma2020pathsrgan} & \multirow{3}{*}{DL} & 31.196 & 0.930 & 0.9407 &	0.0135  & 30.868 & 0.925 & 0.9429  & 0.0115 & 29.24  & \textbf{1.13}  & \textbf{28.93} & \wt{0.077} \\\cline{1-1} \cline{3-14}
			SWD-Net \cite{chen2020joint}&  & \underline{31.368} & \underline{0.937} & 0.9396  & 0.0062 & \textbf{31.033} & \underline{0.933} & 0.9443  & 0.0096 & 3.15  & 50.26  & 1302.34 & \wt{0.129}\\\cline{1-1} \cline{3-14}
			MRC-Net \cite{chen2021super}&  & 31.212 & 0.935 & 0.9561  & \textbf{0.0012} & 30.957 & 0.932 & 0.9450  & 0.0018 & 0.71  & 20.16  & 543.80 & \wt{0.085}  \\\hline
			MRI-RL \cite{li2020mri} & \multirow{3}{*}{RL} & 30.138 & 0.922 & 0.9476  & 0.0080 & 30.735 & 0.928 & 0.9515  & 0.0082 & 0.35  & 12.87  & 92.81 & \wt{0.078}  \\\cline{1-1} \cline{3-14}
			Pixel-RL \cite{furuta2019fully}&  & 30.146 & 0.921 & 0.9470  & 0.0151 & 30.084 & 0.919 & 0.9459  & 0.0145 & \textbf{0.27}  & 9.81  & 73.41 & \wt{0.097} \\\cline{1-1} \cline{3-14}
			\textbf{STAR-RL} &  & \textbf{31.403}  & \textbf{0.939} & \textbf{0.9582}  & \underline{0.0059} & \underline{30.984}  & \textbf{0.935} & \textbf{0.9541}  & \textbf{0.0012}  & \underline{0.50}  & \underline{2.66}  & \underline{70.08} & \underline{\wt{0.061}} \\
			
			\bottomrule[1.1pt]
		\end{tabular}
		\label{sota}
	\end{table*}
	
	\begin{figure*}[!h]
		\centering
		\includegraphics[width=0.9\linewidth]{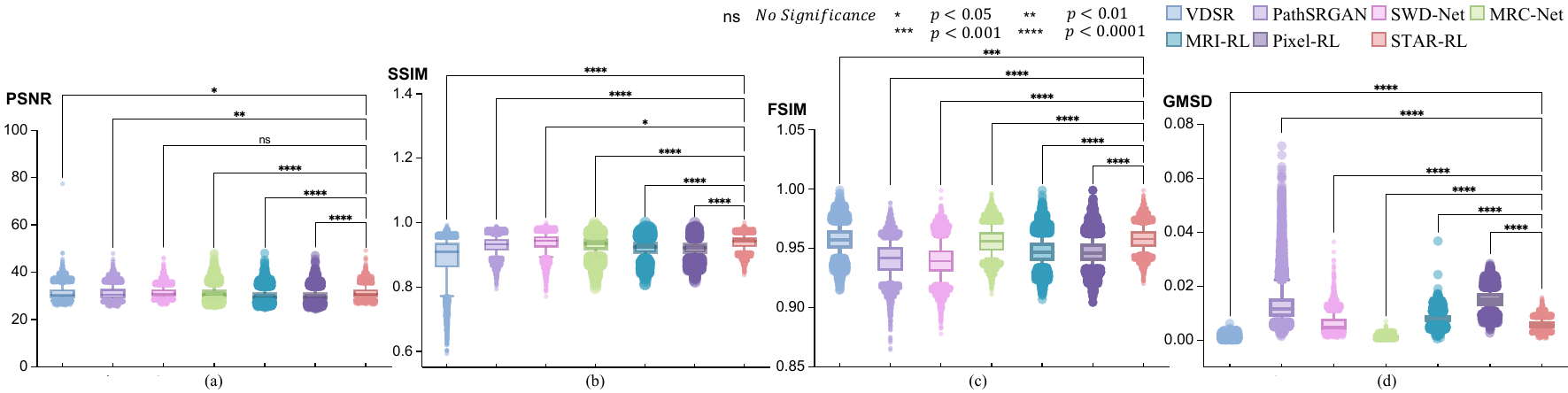}
		\caption{The statistical analysis between STAR-RL and existing SR methods for two evaluation metrics, i.e. (a) SSIM, (b) PSNR, (c) FSIM and (d) GMSD. A box plot is used to visualize the 5th $\sim$ 95th percentiles of results, while deviation outliers are represented as individual points to better observe the distribution.}
		\label{fig::statistics}
	\end{figure*}

	\subsection{Hierarchical Super-Resolution Algorithm}
	As illustrated Algorithm \ref{algorithm}, we present the pseudo-code of STAR-RL training algorithm for pathology image SR. $M$, $N$ and $T$ indicate the total episodes for training model, the maximum iteration number of outer loop and the total time step of inner loop. We update the policy for PW after each outer loop is finished and optimize the policy for spM after all the outer loops are done. 
	
	The inference process is shown in  Algorithm \ref{algorithm2}. In the first outer loop, given a LR image, the higher-level spM outputs a goal map $g_n$ and the higher-level predicts its binary signal $b_n$, which decides whether finish the recovery process. If $b_n$ is less than $0$, we stop the outer loop and regard the current input image as the final result. Otherwise, we continue to restore the selected patch for several inner loops and output the SR image with the recovered patch.
	
	With our hierarchical SR algorithm, we can restore the LR images with interpretable sequential operations, which substantially reduces the computational demand and provides additional interpretability for SR process.

	\begin{figure*}[th]
		\centering
		\includegraphics[width=0.9\textwidth]{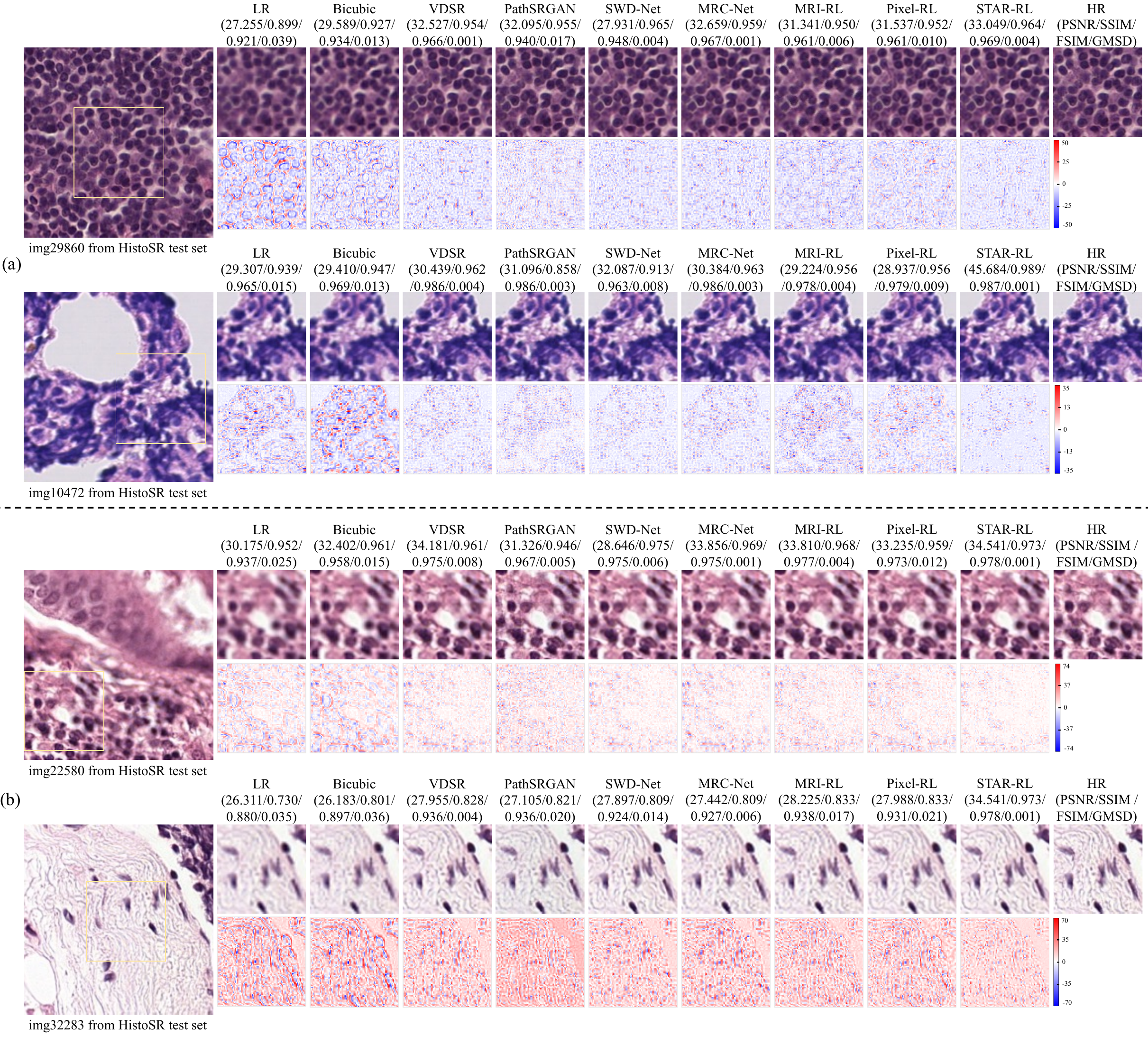}
		\caption{\major{Qualitative comparison of STAR-RL with VDSR \cite{kim2016accurate}, PathSRGAN\cite{ma2020pathsrgan}, SWD-Net \cite{chen2020joint}, MRC-Net \cite{chen2021super}, MRI-RL \cite{li2020mri} and Pixel-RL \cite{furuta2019fully} on HistoSR dataset with (a) BI degradation and (b) GB degradation. In each example, the LR inputs, SR predictions of various SR methods, as well as the HR ground-truth in a yellow box, are shown in the first row. The second row displays the heatmaps of the reconstruction errors,  with white indicating low error and red and blue representing high error, respectively.}} \label{fig::viusalization}
	\end{figure*}

	\section{Experiments}
	\subsection{Experiment Setting} 
	\textbf{Datasets:} The proposed framework is trained and tested on the HistoSR dataset \cite{chen2020joint}, which contains 35,000 images with dimension of $192\times192$. It is divided into three subsets for training (21,000), validation (7,000) and testing (7,000). 
	It includes two versions with different degradation kernels: bicubic (BI) and Gaussian blurring (GB) with $3\times3$ Gaussian kernel ($\sigma=1.0$) followed by bicubic downscaling. Each version comprises low-resolution (LR) and high-resolution (HR) image pairs for training and evaluation.
	
	\textbf{Implementation:} The higher-level spatial manager comprises a backbone model for feature extraction, two policy networks for global horizontal and vertical policies, and a value network for state value estimation. The backbone model uses $Conv3 \times 3$, MaxPool, and another $Conv3 \times 3$ layer sequence. The policy networks have four $MaxPool-Conv1 \times 1$ blocks and a FC layer. The value network has four $Conv3 \times 3$ layers, a $MaxPool$ layer, and two FC layers.
	During training, the higher-level components use the Adam optimizer, while the lower-level policy weights ($\theta_p$) use SGD. The learning rate is initialized as $1\times10^{-4}$ for the higher-level tpM and $1\times10^{-3}$ for other networks, and linearly decayed until 20,000 episodes. In lower-level PW, only $\theta_{\pi_w}$ and $\theta_{V_w}$ networks are optimized during the first 10k episodes, and then these two networks and $\theta_p$ are trained alternatively every 2 episodes. We set the discount factor $\gamma=0.5$ and the predefined patch size $96\times 96$. The batch size is 12. The total time step $T$ for inner loop is set as 3 and maximum iteration number of outer loop is 8. As for the diagnosis network, it is optimized with the Adam optimizer for a maximum of 30 and a batch size of 128. The initial learning rate is set as $1\times10^{-4}$ and decayed linearly. 
	
	\textbf{Evaluation Metrics:} We utilize two standard metrics, Peak Signal-to-Noise Ratio (PSNR) and Structural Similarity Index (SSIM), to evaluate the super-resolution (SR) performance. Additionally, we employ Gradient Magnitude Similarity Deviation (GMSD) and Feature Similarity (FSIM) to evaluate local structure quality. For parameter computation, we count the trainable variables in each network, assuming 32-bit floating-point representation. To calculate FLOPs, we use the number of multiply-add operations, a common practice in deep learning. The memory refers to the cumulative output feature maps from operations like Convolution, ReLU, and Batch Normalization.
	
	\begin{table}[tp]
		\renewcommand{\arraystretch}{1.5}   
		\caption{Ablation studies on HistoSR dataset.}
		\centering
		\label{tab_ablation}
		\scalebox{0.95}{
			\begin{tabular}{ccccccc}
				\toprule[1pt]
				PW & spM & tpM & PSNR  $\uparrow$ & SSIM  $\uparrow$ & FSIM  $\uparrow$ & GMSD  $\downarrow$ \\ \hline 
				&  &  & 30.1385 & 0.9216 & 0.9476 & 0.0080  \\ \hline 
				\checkmark  &  &  & 30.6542 & 0.9309 & 0.9514 & 0.0083\\ \hline 
				\checkmark  & \checkmark & &31.3055 & 0.9375 & 0.9580 & 0.0063  \\\hline
				\checkmark  & \checkmark  & \checkmark  &\textbf{31.4032} & \textbf{0.9388} & \textbf{0.9582} & \textbf{0.0059}  \\ \hline
				\toprule[1pt]
		\end{tabular}}
	\end{table}
	
	\begin{table}[tp]
		\centering
		\caption{\major{The effectiveness of different patch sizes.}}
		\scalebox{0.8}{
			\renewcommand{\arraystretch}{1.4}
			\begin{tabular}{cccccccc}
				\toprule[1pt]
				\multirow{2}{*}{Patch size} & \multirow{2}{*}{PSNR  $\uparrow$} & \multirow{2}{*}{SSIM  $\uparrow$} & \multirow{2}{*}{FSIM  $\uparrow$} & \multirow{2}{*}{GMSD  $\downarrow$} & \multirow{2}{*}{\makecell[c]{Params \\(MB)}}& \multirow{2}{*}{\makecell[c]{FLOPs\\(GB)}} & \multirow{2}{*}{\makecell[c]{Memory\\(MB)}} \\ 
				& & & & &    &    &   \\ \hline
				32 & 30.061 & 0.920 & 0.9435 & 0.0126 & 0.515 & 0.46 & 53.58\\ \hline
				48 & 30.399 & 0.925 & 0.9456 & 0.0111 & 0.512 & 0.80 & 56.16\\ \hline
				64 & 30.755 & 0.930 & 0.9505 & 0.0097 & 0.508 & 1.28 & 59.77\\ \hline
				96 & \textbf{31.403} & \textbf{0.939 }& \textbf{0.9582} & \textbf{0.0059} & 0.502 & 2.66 & 70.08\\ \hline
				120 & 31.125 & 0.936 & 0.9556 & 0.0084 & 0.530 & 4.41 & 150.04\\ \hline
				144 & 31.028 & 0.934 & 0.9540 & 0.0084 & 0.515 & 6.12 & 162.81 \\ \hline
				168 & 31.384 & 0.939 & 0.9581 & 0.0076 & 0.502 & 8.14 & 177.89 \\ 
				\toprule[1pt]
			\end{tabular}
		}
		\label{tab::diff_patch_size}
	\end{table}
	
	\begin{table*}[th]
		\renewcommand{\arraystretch}{1.3}   
		\centering
		\caption{The generalization ability of the proposed method on HistoSR dataset with different GB degradation ($\sigma=\left \{ 0.6,0.8,1.0,1.2,1.4 \right \}$). The best results are \textbf{highlighted}.}
		\begin{tabular}{c|c|c|c|c|c|c|c|c|c|c}
			\toprule[1pt]
			\multirow{2}{*}{Methods} &\multicolumn{2}{c|}{$\sigma=0.6$}  & \multicolumn{2}{c|}{$\sigma=0.8$}  & \multicolumn{2}{c|}{$\sigma=1.0$} & \multicolumn{2}{c|}{$\sigma=1.2$} & \multicolumn{2}{c}{$\sigma=1.4$} \\ \cline{2-11}
			& SSIM $\uparrow$ & FSIM $\uparrow$ & SSIM $\uparrow$ & FSIM $\uparrow$ &  SSIM $\uparrow$ & FSIM $\uparrow$ &  SSIM $\uparrow$ & FSIM $\uparrow$ &  SSIM $\uparrow$ & FSIM $\uparrow$\\ \hline
			
			VDSR \cite{kim2016accurate} & 0.909  & 0.939  & 0.893  & 0.926  & 0.895  & 0.954  & 0.876  & 0.914  & 0.873  & 0.911 \\ \hline
			PathSRGAN \cite{ma2020pathsrgan} & 0.893  & 0.935  & 0.901  & 0.942  & 0.925  & 0.943  & 0.902  & 0.943  & 0.901  & 0.943 \\ \hline
			SWD-Net \cite{chen2020joint} & 0.917  & 0.944  & 0.898  & 0.930  & 0.933  & 0.944  & 0.881  & 0.917  & 0.876  & 0.914 \\ \hline
			MRC-Net \cite{chen2021super} & 0.899  & 0.931  & 0.884  & 0.920  & 0.932  & 0.945  & 0.869  & 0.909  & 0.866  & 0.906 \\ \hline
			MRI-RL \cite{li2020mri} & 0.917  & 0.947  & 0.922  & 0.952  & 0.928  & 0.951  & 0.921  & 0.951  & 0.919  & 0.950 \\ \hline
			Pixel-RL \cite{furuta2019fully} & 0.905  & 0.943  & 0.912  & 0.946  & 0.919  & 0.946  & 0.913  & 0.945  & 0.913  & 0.945 \\ \hline
			STAR-RL & \textbf{0.927}  & \textbf{0.952}  & \textbf{0.930}  & \textbf{0.955}  & \textbf{0.935}  & \textbf{0.954}  & \textbf{0.928}  & \textbf{0.953}  & \textbf{0.927}  & \textbf{0.952} \\
			\bottomrule[1.1pt]
		\end{tabular}
		\label{generalization}
	\end{table*}
	
	\begin{table}[h]
		\centering
		\caption{The analysis on different degradation types with bicubic downsampling. BI represents bicubic interpolation. Asterisks indicate statistical significance: n.s. No Significance, * p $<$ 0.05, ** p $<$ 0.01, *** p $<$ 0.001.}
		\renewcommand{\arraystretch}{1.4}
		\scalebox{0.85}{
			\begin{tabular}{c|c|c|c|c|c}
				\toprule[1pt]
				Degradation & Methods & PSNR $\uparrow$ & SSIM $\uparrow$ & FSIM $\uparrow$  & GMSD $\downarrow$ \\ \hline
				\multirow{3}{*}{/} &  BI & ${28.847}^{***}$ &
				${0.897}^{***}$ & ${0.9225}^{***}$ & ${0.0158}^{***}$  \\ 
				& SWD-Net & ${31.368}^{*}$ & ${0.937}^{**}$ & ${0.9396}^{***}$ & ${0.0062}^{*}$ \\ 
				& STAR-RL & 31.403  & 0.939  & 0.9582  & 0.0059 \\ \hline
				\multirow{3}{*}{Blur} & BI & ${28.179}^{**}$ & ${0.881}^{***}$ & ${0.9101}^{**}$ & ${0.0229}^{***}$ \\ 
				& SWD-Net & ${31.033}^{*}$ & ${0.933}^{n.s.}$ & ${0.9443}^{***}$ & ${0.0096}^{***}$ \\ 
				& STAR-RL & 30.984  & 0.935  & 0.9541  & 0.0012\\ \hline
				\multirow{3}{*}{Pepper} & BI & ${27.198}^{***}$ & ${0.879}^{***}$ & ${0.9180}^{***}$ & ${0.0301}^{**}$  \\
				& SWD-Net & ${31.385}^{n.s.}$ & ${0.933}^{*}$ & ${0.9575}^{**}$ & ${0.0071}^{**}$  \\ 
				& STAR-RL & 31.457  & 0.939  & 0.9592  & 0.0061   \\ \hline
				\multirow{3}{*}{Salt} & BI & ${28.303}^{**}$ & ${0.890}^{***}$ & ${0.9210}^{***}$ & ${0.0192}^{**}$   \\ 
				& SWD-Net & ${30.810}^{***}$ & ${0.926}^{***}$ & ${0.9526}^{*}$ & ${0.0083}^{***}$   \\ 
				& STAR-RL & 30.920  & 0.933  & 0.9557  & 0.0074  \\ \hline
				\multirow{3}{*}{Gaussian} & BI &${28.381}^{**}$ & ${0.879}^{***}$ & ${0.9207}^{**}$ & ${0.0167}^{**}$  \\ 
				& SWD-Net & ${30.579}^{***}$ & ${0.922}^{***}$ & ${0.9503}^{*}$ & ${0.0094}^{***}$   \\ 
				& STAR-RL & 30.900  & 0.930  & 0.9541  & 0.0082 \\ \hline
				\toprule[1pt]
		\end{tabular}}
		\label{tab::degrad_eval_NCT}
	\end{table}

	\begin{table}[h]
		\centering
		\caption{\major{The SR performance of different image sizes.}}
		\renewcommand{\arraystretch}{1.4}
		\begin{tabular}{ccccc}
			\toprule[1pt]
			Image size & PSNR $\uparrow$ & SSIM $\uparrow$ & FSIM $\uparrow$ & GMSD $\downarrow$ \\ \hline
			96 & 32.3700  & 0.9428  & 0.9611  & 0.0043  \\ \hline
			192 & 31.4030  & 0.9390  & 0.9582  & 0.0059  \\ \hline
			288 & 31.6882  & 0.9367  & 0.9558  & 0.0046  \\ \hline
			384 & 31.2619  & 0.9351  & 0.9544  & 0.0052  \\ \hline
			\toprule[1pt]
		\end{tabular}
		\label{tab::diff_img_size}
	\end{table}

	\subsection{Super Resolution Results}
	\subsubsection{Quantitative Results}
	Table \ref{sota} lists the SR performance comparison between the STAR-RL and existing SR networks based on deep learning (DL) \cite{kim2016accurate,ma2020pathsrgan,chen2020joint,chen2021super} and reinforcement learning (RL) \cite{li2020mri,furuta2019fully}. 
	Among the compared research works, the PathSRGAN \cite{ma2020pathsrgan}, SWD-Net \cite{chen2020joint}, MRC-Net \cite{chen2021super} and MRI-RL \cite{li2020mri} are SR networks for medical images, {while the VDSR \cite{kim2016accurate} and  Pixel-RL \cite{furuta2019fully} work for natural images.} For the BI degradation model, when comparing with RL based methods, STAR-RL surpasses the MRI-RL \cite{li2020mri} and Pixel-RL \cite{furuta2019fully}, with significant improvement of 1.265 dB and 1.257 dB respectively, indicating its superiority for global recover capacity over the existing SR networks. 
	{Moreover, the proposed STAR-RL achieves the best performance in metrics for local structures, with 0.9541 in FSIM and 0.0012 in GMSD. This suggests that the SR images generated by our method preserve better biological details than those generated by existing methods.} {In addition, the STAR-RL achieves comprehensive abilities with 0.5 MB params, 2.66 GB FLOPs and 70.08 MB memory, indicating the effective design of hierarchical architecture for computational efficiency.}
	
	To observe the statistical properties of the results, we conduct the statistical analysis between the proposed STAR-RL and existing SR methods on the HistoSR dataset with BI degradation, as shown in Fig. \ref{fig::statistics}. Apparently, STAR-RL has fewer deviation outliers and the distribution is concentrated with smaller variance compared to existing methods. The significant difference between STAR-RL and other methods also can be observed. These reveal the superiority of the proposed STAR-RL to the existing SR methods in terms of SSIM, PSNR, FSIM and GMSD metrics statistically.
	
	\subsubsection{Qualitative Results}
	{To qualitatively evaluate the performance on the HistoSR dataset with BI and GB degradation, we compare the SR images among the existing methods as shown in Fig. \ref{fig::viusalization}.}
	The visual results show that the proposed STAR-RL synthesizes the high-quality reconstructed details with clear structure information similar to the HR image. To further compare with existing methods, we visualized the heatmap of reconstruction errors, which were normalized by the maximum and minimum reconstruction errors of different methods on the same sample. Specifically, STAR-RL reduces reconstruction errors substantially within the marked region, whose heatmaps preserve more white areas and fewer red and blue areas. This suggests the superiority of the proposed STAR-RL to existing SR methods.
	
	\begin{figure*}[t]
		\centering
		\includegraphics[width=0.85\textwidth]{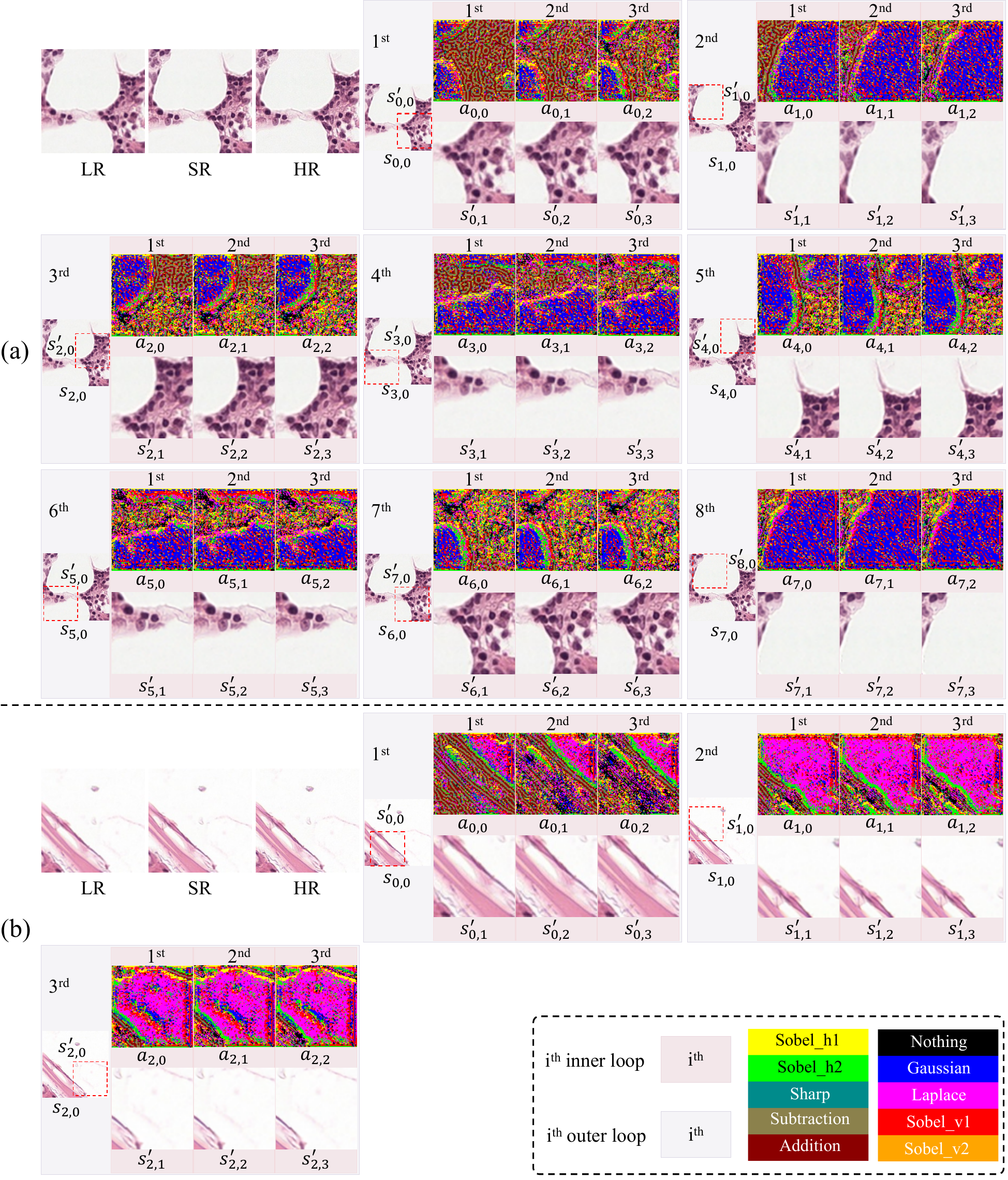}
		\caption{\major{Visualization of the outer loops and the inner loops for two examples. In the outer loop, the higher-level spM predicts the goal map to select the most corrupted patches, i.e. state $s_{n,0}^{'}$. In the inner loop, the lower-level PW restores the selected patches $s_{n,0}^{'}$ to different states $s_{n,t}^{'}(t= \left \{1,2,3 \right \})$ with different interpretable actions $a_{n,t}$ sequentially. The predefined interpretable actions are depicted in different colors. We do not show the outer loop if the higher-level tpM determines to stop earlier.}}
		\label{fig::vis_actions}
	\end{figure*}
	
	\subsection{Ablation Study}
	To verify the effectiveness of each module, detailed ablation studies have been conducted on the HistoSR dataset with BI degradation. 
	As illustrated in Table \ref{tab_ablation}, when integrating the higher-level spM with the lower-level PW, the performance is remarkably increased by 1.167 dB of PSNR and 0.38\% of FSIM over the baseline. In addition, the tpM improves the PSNR by 0.098 dB.
	These results demonstrate that we can improve the overall efficiency and effectiveness of image super-resolution by prioritizing computational resources and carefully controlling the recovery process.
	
	\change{In addition, we designed an ablation study about the influence of patch size on the SR performance as shown in Table \ref{tab::diff_patch_size}. When the patch size was smaller than 96, the results clearly indicate a correlation between patch size and super-resolution quality, with performance metrics declining as the patch size decreases. We hypothesized that this occurs due to the reduced patch size, affecting the perception of objects within each patch and consequently altering the action types selected by the algorithm. When the patch size increased beyond 96, the SR performance was comparable to that of the 96 patch size. However, larger patch sizes would inevitably increase resource consumption. Considering both resource utilization and performance, we adopte a patch size of 96. }

	\begin{table}[t]
		\renewcommand{\arraystretch}{1.5}   
		\caption{The promotion on tumor recognition task (\%). P-value are calculated, respectively, of the corresponding method and the STAR-RL. Asterisks indicate statistical significance: n.s. No Significance, * p $<$ 0.05, ** p $<$ 0.01, *** p $<$ 0.001.}
		\centering
		\label{tab_tumor}
		\begin{tabular}{c|c|c}
			\toprule[1pt]
			
			Methods & Accuracy & F1 score  \\ \hline
			Origin & ${93.06 \pm 1.01 }^{***}$& ${68.91 \pm 1.41}^{***}$   \\ \hline
			Bicubic & ${94.08 \pm 1.28}^{***}$ & ${69.98 \pm 1.70}^{***}$   \\ \hline
			VDSR    & ${96.06 \pm 2.08}^{**}$ & ${72.03 \pm 2.70}^{*}$   \\ \hline 
			PathSRGAN & ${96.04 \pm 1.79}^{n.s.}$ & ${72.09 \pm 2.25}^{n.s.}$  \\ \hline
			SWD-Net  & ${96.26 \pm 1.52}^{**}$ & ${72.24 \pm 2.22}^{***}$   \\ \hline
			MRC-Net  & ${96.18 \pm 1.90}^{***}$ & ${72.25 \pm 2.39}^{**}$   \\ \hline
			MRI-RL & ${96.13 \pm 1.81}^{***}$ & ${72.13 \pm 2.30}^{***}$   \\ \hline
			Pixel-RL & ${96.10 \pm 1.88}^{**}$ & ${72.03 \pm 2.49}^{***}$ \\ \hline
			\textbf{STAR-RL} & \textbf{96.41±1.94} & \textbf{72.45±2.48}  \\ \hline
			\toprule[1pt]
		\end{tabular}
	\end{table}

	\subsection{Interpretability}
	To unveil the interpretability of STAR-RL, we visualize reconstruction process for two examples. As shown in Fig. \ref{fig::vis_actions} (a), we visualize eight outer loops. In the first outer loop, the higher-level spM selects the most corrupted regions $s_{0,0}^{'}$ in $s_{0,0}$, where $s_{0,0}^{'}$ covers many tissue cells. When recovering the selected patches in the first inner loop, the addition and subtraction are used to strengthen pixels for the edges and pixels outside the edges. In the next two inner loops, Sobel filters are adopted to further refine the edges, the Laplace and Gaussian filters are utilized to smooth the over-sharpened edges and restore the black spaces. Apparently, the recovered patches $s_{0,t}^{'}(t={1,2,3})$ preserve more details with sharper edges and approaches to the HR one.
	These indicate that the reconstruction process can be decomposed to several visible sequential decision problems and interpreted by the predefined interpretable actions. This interpretable process avoids unexpected pattern generation. In example (b), we display three outer loops, since the higher-level tpM requires to stop in the fourth outer loop. It suggests that the proposed STAR-RL is efficient to restore the LR image with less computational time and remain the SR performance.
	
	\section{Discussion}
	
	\subsection{Generalizability}
	Models that generalize well are more likely to be applicable in different clinical settings \cite{guo2021semantic}. To evaluate model generalizability in various degradation, we conduct several experiments including varying the sigma of Gaussian blur kernel, adding more additional types of noise and executing inference on image with different sizes.
	We train the model on LR image with $\sigma=1.0$ Gaussian kernel and test our method on LR image with $\sigma=\left \{0.6,0.8,1.0,1.2,1.4\right \}$ Gaussian kernel. As listed in Table \ref{generalization}, the STAR-RL achieves the best SR performance under different $\theta$, surpassing other methods by a substantial margin in SSIM and FSIM. It suggests that STAR-RL shows the best generalized ability under different degradation and powerful practicability in real-world scenarios.
	Also, we have expanded our experimental validation to include additional types of noise—namely, blur, pepper noise, salt noise, and Gaussian noise—in conjunction with bicubic downsampling. These experiments were designed to simulate a broader range of potential real-world degradation scenarios. The results of these tests are included in the Table \ref{tab::degrad_eval_NCT}. The STAR-RL framework exhibits satisfactory capabilities in recovering from these additional degradations, thereby demonstrating its effectiveness in handling a variety of noise conditions.
	In addition, we further conduct super-resolution experiments on images with different sizes 96, 192, 288, and 384 to demonstrate the flexibility in processing images of different scales, as shown in Table \ref{tab::diff_img_size}. It is noted that our framework is built upon a fully convolutional network architecture, which inherently supports input images of varying sizes without the need for resizing to a standard dimension.
	
	\subsection{Disease Diagnosis Enhancement}
	In order to investigate the promotion of SR methods on disease diagnosis tasks, experiments for tumor recognition are carried out on the PCam dataset \cite{veeling2018rotation}. \major{We have split the data into a 3:1:1 ratio for training, validation, and testing with a strict patient-level split. Additionally, we have conducted cross-validation and calculated the P-value to validate the statistical significance of our model performance. We have adopted the SR networks as pre-processing for the input image and the pre-trained ResNet-18 \cite{he2016deep} as a classifier. 
		Table \ref{tab_tumor} shows that the proposed STAR-RL performs significantly better than other methods and achieves optimal performance with an accuracy of 96.41\%. It manifests that the STAR-RL can help the diagnosis task by generating more discriminative features.}
	
	\subsection{Experiment on Gigapixel Images}
\begin{figure}[t]
		\centering
		\includegraphics[width=\linewidth]{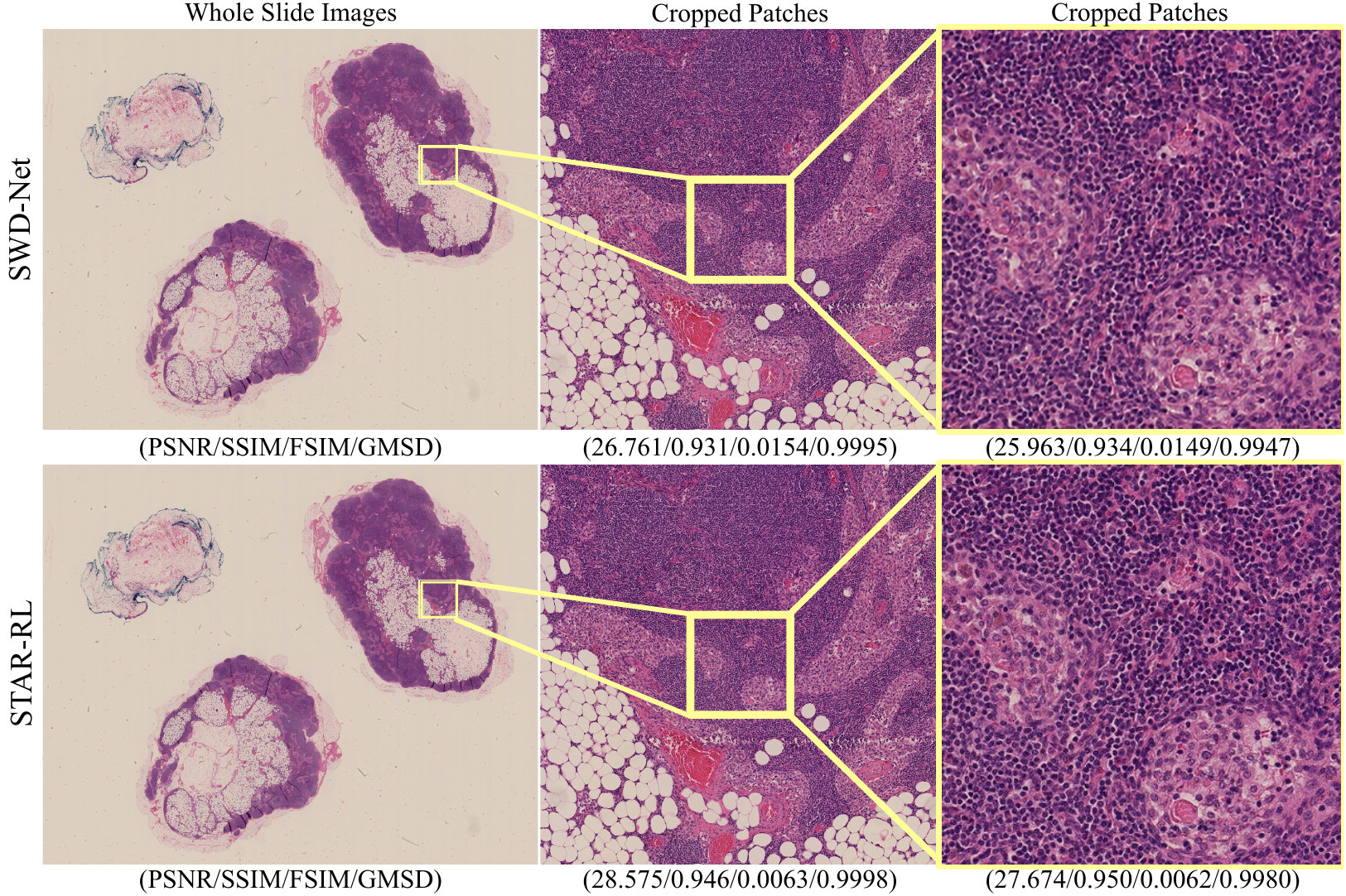}
		\caption{Visual results of the SR images generated by STAR-RL and SWD-Net.}
		\label{fig::wsi_comp}
	\end{figure}

	\change{To validate our method's effectiveness on whole-slide images (WSIs), we conducted experiments applying STAR-RL and other methods to gigapixel images from the Camelyon16 dataset, representative of true WSI complexity. As shown in Table~\ref{tab::WSI_SR}, STAR-RL achieves the best WSI SR performance, outperforming current methods with statistical robustness demonstrated by the p-values. Figure \ref{fig::wsi_comp} shows a recovered WSI from STAR-RL and the second-best SWD-Net, where STAR-RL exhibits higher gigapixel-level performance. This enhancement stems from STAR-RL's strategic computational resource allocation, devoting more resources to non-flat regions and fewer to flat regions, enabling robust WSI resolution enhancement crucial for diagnostic clarity and detail. While our approach can reconstruct visually superior images, it may not reveal additional diagnostic information beyond the input data. We emphasize that conventional multi-level whole slide imaging, collecting higher-resolution regions of interest detected at each level, remains necessary to unveil finer biomedically relevant details.}

	\begin{table}[h]
		
		\centering
		\caption{\change{The WSI SR performance comparison with different methods. P-value are calculated, respectively, of corresponding method and the STAR-RL. Asterisks indicate statistical significance: n.s. No Significance, * p $<$ 0.05, ** p $<$ 0.01, *** p $<$ 0.001.} }
		\renewcommand{\arraystretch}{1.4}
		\begin{tabular}{ccccc}
			\toprule[1pt]
			Methods & PSNR $\uparrow$ & SSIM $\uparrow$& FSIM $\uparrow$& GMSD $\downarrow$\\ \hline
			VDSR & ${31.3539}^{***}$ & ${0.8865}^{***}$ & ${0.9574}^{***}$ & ${0.0012}^{***}$\\
			PathSRGAN & \underline{${31.2388}^{**}$} & ${0.9296}^{***}$ & ${0.7899}^{***}$ & ${0.1348}^{***}$\\
			SWD-Net & ${31.3200}^{n.s.}$ & \underline{${0.9368}^{n.s.}$} & ${0.9396}^{***}$ & $\mathbf{{0.0062}^{***}}$\\
			MRC-Net & ${31.1880}^{***}$ & ${0.9308}^{***}$ & \underline{${0.9561}^{***}$} & ${0.0012}^{***}$\\
			MRI-RL & ${30.1304}^{***}$ & ${0.9216}^{***}$ & ${0.9476}^{***}$ & ${0.0080}^{***}$\\
			Pixel-RL & ${30.1377}^{***}$ & ${0.9211}^{***}$ & ${0.9470}^{***}$ & ${0.0151}^{***}$\\
			STAR-RL & \textbf{31.3951} & \textbf{0.9388} &   \textbf{0.9582}  &   \underline{0.0070}   \\ 
			
			\toprule[1pt]
		\end{tabular}
		\label{tab::WSI_SR}
	\end{table}
	
	\subsection{Limitation}
	The proposed STAR-RL focuses on local enhancement operations, which may be inadequate for addressing global degradations like image-wide artifacts or non-local degradation patterns. Generative models could have an advantage here by learning high-level representations of entire images, potentially restoring coherence across large areas more effectively than local filters. Our future work could explore integrating non-local processing strategies \cite{dabov2007image,gu2014weighted} within STAR-RL, leveraging self-similarity across scales and regions to inform enhancement operations. \change{Moreover, while this work recovers HR images from their LR counterparts using reinforcement learning to select the optimal sequence of operations, our current approach still requires ground truth HR images to learn the appropriate operations for reconstructing HR patches from LR patches, aligning with supervised learning. In future work, we aim to explore mining information directly from the image itself to enable reinforcement learning without relying on paired LR-HR datasets. This approach could lead to more data-efficient and generalizable super-resolution models, circumventing the need for such datasets.}

	\section{Conclusions}
	In this work, we present a novel pathology image super-resolution framework named Spatial-Temporal hierARchical Reinforcement Learning (STAR-RL) for medical image super-resolution. 
	To facilitate the interpretability of recovery process, we reformulate the SR problem as a Markov decision problem of interpretable operations. 
	We further propose a two-level hierarchical mechanism including higher-level spatial Manager (spM), temporal Manager (tpM) and lower-level Patch Worker (PW). 
	Apart from global evaluation metrics PSNR and SSIM, we adopt the FSIM and GMSD to assess the quality of local structures, i.e. biomedical details.
	Extensive experiments demonstrate the advantages of STAR-RL framework over other SR methods in both global and local metrics. Moreover, we analyze the interpretable recovery process, which avoids unforeseen pattern generation. Furthermore, the STAR-RL framework shows strong generalizability in various degraded scenarios, as well as promotion on diagnosis tasks.

	\bibliographystyle{IEEEtran}
	\bibliography{reference}

\begin{thebibliography}{10}
\providecommand{\url}[1]{#1}
\csname url@samestyle\endcsname
\providecommand{\newblock}{\relax}
\providecommand{\bibinfo}[2]{#2}
\providecommand{\BIBentrySTDinterwordspacing}{\spaceskip=0pt\relax}
\providecommand{\BIBentryALTinterwordstretchfactor}{4}
\providecommand{\BIBentryALTinterwordspacing}{\spaceskip=\fontdimen2\font plus
\BIBentryALTinterwordstretchfactor\fontdimen3\font minus
  \fontdimen4\font\relax}
\providecommand{\BIBforeignlanguage}[2]{{%
\expandafter\ifx\csname l@#1\endcsname\relax
\typeout{** WARNING: IEEEtran.bst: No hyphenation pattern has been}%
\typeout{** loaded for the language `#1'. Using the pattern for}%
\typeout{** the default language instead.}%
\else
\language=\csname l@#1\endcsname
\fi
#2}}
\providecommand{\BIBdecl}{\relax}
\BIBdecl

\bibitem{ma2020pathsrgan}
J.~Ma, J.~Yu, S.~Liu, L.~Chen, X.~Li, J.~Feng, Z.~Chen, S.~Zeng, X.~Liu, and
  S.~Cheng, ``Pathsrgan: Multi-supervised super-resolution for cytopathological
  images using generative adversarial network,'' \emph{IEEE Trans. Med. Imag.},
  vol.~39, no.~9, pp. 2920--2930, 2020.

\bibitem{chen2020joint}
Z.~Chen, X.~Guo, C.~Yang, B.~Ibragimov, and Y.~Yuan, ``Joint spatial-wavelet
  dual-stream network for super-resolution,'' in \emph{MICCAI}.\hskip 1em plus
  0.5em minus 0.4em\relax Springer, 2020, pp. 184--193.

\bibitem{cornish2012whole}
T.~C. Cornish, R.~E. Swapp, and K.~J. Kaplan, ``Whole-slide imaging: routine
  pathologic diagnosis,'' \emph{Adv. Anat. Pathol.}, vol.~19, no.~3, pp.
  152--159, 2012.

\bibitem{pantanowitz2011review}
L.~Pantanowitz, P.~N. Valenstein, A.~J. Evans, K.~J. Kaplan, J.~D. Pfeifer,
  D.~C. Wilbur, L.~C. Collins, and T.~J. Colgan, ``Review of the current state
  of whole slide imaging in pathology,'' \emph{J. Pathol. Inform.}, vol.~2,
  no.~1, p.~36, 2011.

\bibitem{afshari2023single}
M.~Afshari, S.~Yasir, G.~L. Keeney, R.~E. Jimenez, J.~J. Garcia, and H.~R.
  Tizhoosh, ``Single patch super-resolution of histopathology whole slide
  images: a comparative study,'' \emph{J. Med. Imaging}, vol.~10, no.~1, p.
  017501, 2023.

\bibitem{ghaznavi2013digital}
F.~Ghaznavi, A.~Evans, A.~Madabhushi, and M.~Feldman, ``Digital imaging in
  pathology: whole-slide imaging and beyond,'' \emph{Annu. Rev. Pathol.: Mech.
  Dis.}, vol.~8, pp. 331--359, 2013.

\bibitem{park2019recent}
J.~Park, Y.~Hwang, J.-H. Yoon, M.-G. Park, J.~Kim, Y.~J. Lim, and H.~J. Chun,
  ``Recent development of computer vision technology to improve capsule
  endoscopy,'' \emph{Clin. Endosc.}, vol.~52, no.~4, p. 328, 2019.

\bibitem{madabhushi2016image}
A.~Madabhushi and G.~Lee, ``Image analysis and machine learning in digital
  pathology: Challenges and opportunities,'' \emph{Med. Image Anal.}, vol.~33,
  pp. 170--175, 2016.

\bibitem{srivastav2019human}
V.~Srivastav, A.~Gangi, and N.~Padoy, ``Human pose estimation on
  privacy-preserving low-resolution depth images,'' in \emph{MICCAI}.\hskip 1em
  plus 0.5em minus 0.4em\relax Springer, 2019, pp. 583--591.

\bibitem{liu2022edge}
Y.~Liu, J.~Liu, and Y.~Yuan, ``Edge-oriented point-cloud transformer for 3d
  intracranial aneurysm segmentation,'' in \emph{International Conference on
  Medical Image Computing and Computer-Assisted Intervention}.\hskip 1em plus
  0.5em minus 0.4em\relax Springer, 2022, pp. 97--106.

\bibitem{chang2004super}
H.~Chang, D.-Y. Yeung, and Y.~Xiong, ``Super-resolution through neighbor
  embedding,'' in \emph{CVPR}, vol.~1.\hskip 1em plus 0.5em minus 0.4em\relax
  IEEE, 2004, pp. I--I.

\bibitem{su2005neighborhood}
K.~Su, Q.~Tian, Q.~Xue, N.~Sebe, and J.~Ma, ``Neighborhood issue in
  single-frame image super-resolution,'' in \emph{ICME}.\hskip 1em plus 0.5em
  minus 0.4em\relax IEEE, 2005, pp. 4 pp.--.

\bibitem{kim2010single}
K.~I. Kim and Y.~Kwon, ``Single-image super-resolution using sparse regression
  and natural image prior,'' \emph{IEEE Trans. Pattern Anal. Mach. Intell.},
  vol.~32, no.~6, pp. 1127--1133, 2010.

\bibitem{sun2008image}
J.~Sun, Z.~Xu, and H.-Y. Shum, ``Image super-resolution using gradient profile
  prior,'' in \emph{CVPR}.\hskip 1em plus 0.5em minus 0.4em\relax IEEE, 2008,
  pp. 1--8.

\bibitem{timofte2013anchored}
R.~Timofte, V.~De~Smet, and L.~Van~Gool, ``Anchored neighborhood regression for
  fast example-based super-resolution,'' in \emph{ICCV}, 2013, pp. 1920--1927.

\bibitem{kim2016accurate}
J.~Kim, J.~K. Lee, and K.~M. Lee, ``Accurate image super-resolution using very
  deep convolutional networks,'' in \emph{CVPR}, 2016, pp. 1646--1654.

\bibitem{tian2020coarse}
C.~Tian, Y.~Xu, W.~Zuo, B.~Zhang, L.~Fei, and C.-W. Lin, ``Coarse-to-fine cnn
  for image super-resolution,'' \emph{IEEE Trans. Multimed.}, vol.~23, pp.
  1489--1502, 2020.

\bibitem{liu2020residual}
J.~Liu, W.~Zhang, Y.~Tang, J.~Tang, and G.~Wu, ``Residual feature aggregation
  network for image super-resolution,'' in \emph{CVPR}, 2020, pp. 2359--2368.

\bibitem{zhou2021ultrasound}
Z.~Zhou, Y.~Guo, and Y.~Wang, ``Ultrasound deep beamforming using a
  multiconstrained hybrid generative adversarial network,'' \emph{Med. Image
  Anal.}, vol.~71, p. 102086, 2021.

\bibitem{liang2022details}
J.~Liang, H.~Zeng, and L.~Zhang, ``Details or artifacts: A locally
  discriminative learning approach to realistic image super-resolution,'' in
  \emph{CVPR}, 2022, pp. 5657--5666.

\bibitem{liang2021hierarchical}
J.~Liang, A.~Lugmayr, K.~Zhang, M.~Danelljan, L.~Van~Gool, and R.~Timofte,
  ``Hierarchical conditional flow: A unified framework for image
  super-resolution and image rescaling,'' in \emph{ICCV}, 2021, pp. 4076--4085.

\bibitem{mukherjee2019super}
L.~Mukherjee, H.~D. Bui, A.~Keikhosravi, A.~Loeffler, and K.~W. Eliceiri,
  ``Super-resolution recurrent convolutional neural networks for learning with
  multi-resolution whole slide images,'' \emph{J. Biomed. Opt.}, vol.~24,
  no.~12, p. 126003, 2019.

\bibitem{mukherjee2018convolutional}
L.~Mukherjee, A.~Keikhosravi, D.~Bui, and K.~W. Eliceiri, ``Convolutional
  neural networks for whole slide image superresolution,'' \emph{Biomed. Opt.
  Express}, vol.~9, no.~11, pp. 5368--5386, 2018.

\bibitem{chen2021super}
Z.~Chen, X.~Guo, P.~Y. Woo, and Y.~Yuan, ``Super-resolution enhanced medical
  image diagnosis with sample affinity interaction,'' \emph{IEEE Trans. Med.
  Imag.}, vol.~40, no.~5, pp. 1377--1389, 2021.

\bibitem{li2021single}
B.~Li, A.~Keikhosravi, A.~G. Loeffler, and K.~W. Eliceiri, ``Single image
  super-resolution for whole slide image using convolutional neural networks
  and self-supervised color normalization,'' \emph{Med. Image Anal.}, vol.~68,
  p. 101938, 2021.

\bibitem{riddle2011plasticity}
N.~C. Riddle, A.~Minoda, P.~V. Kharchenko, A.~A. Alekseyenko, Y.~B. Schwartz,
  M.~Y. Tolstorukov, A.~A. Gorchakov, J.~D. Jaffe, C.~Kennedy, D.~Linder-Basso
  \emph{et~al.}, ``Plasticity in patterns of histone modifications and
  chromosomal proteins in drosophila heterochromatin,'' \emph{Genome Res.},
  vol.~21, no.~2, pp. 147--163, 2011.

\bibitem{donovan2021mitotic}
T.~A. Donovan, F.~M. Moore, C.~A. Bertram, R.~Luong, P.~Bolfa, R.~Klopfleisch,
  H.~Tvedten, E.~N. Salas, D.~B. Whitley, M.~Aubreville \emph{et~al.},
  ``Mitotic figures—normal, atypical, and imposters: a guide to
  identification,'' \emph{Vet. Pathol.}, vol.~58, no.~2, pp. 243--257, 2021.

\bibitem{xue2013gradient}
W.~Xue, L.~Zhang, X.~Mou, and A.~C. Bovik, ``Gradient magnitude similarity
  deviation: A highly efficient perceptual image quality index,'' \emph{IEEE
  Trans. on Image Process.}, vol.~23, no.~2, pp. 684--695, 2013.

\bibitem{zhang2011FSIM}
L.~Zhang, L.~Zhang, X.~Mou, and D.~Zhang, ``Fsim: A feature similarity index
  for image quality assessment,'' \emph{IEEE Trans. on Image Process.},
  vol.~20, no.~8, pp. 2378--2386, 2011.

\bibitem{kanavati2022deep}
F.~Kanavati, N.~Hirose, T.~Ishii, A.~Fukuda, S.~Ichihara, and M.~Tsuneki, ``A
  deep learning model for cervical cancer screening on liquid-based cytology
  specimens in whole slide images,'' \emph{Cancers}, vol.~14, no.~5, p. 1159,
  2022.

\bibitem{schapiro2022mcmicro}
D.~Schapiro, A.~Sokolov, C.~Yapp, Y.-A. Chen, J.~L. Muhlich, J.~Hess, A.~L.
  Creason, A.~J. Nirmal, G.~J. Baker, M.~K. Nariya \emph{et~al.}, ``Mcmicro: a
  scalable, modular image-processing pipeline for multiplexed tissue imaging,''
  \emph{Nat. Methods}, vol.~19, no.~3, pp. 311--315, 2022.

\bibitem{gonzalez2009digital}
R.~C. Gonzalez, \emph{Digital image processing}.\hskip 1em plus 0.5em minus
  0.4em\relax Pearson education india, 2009.

\bibitem{chen2022dynamic}
W.~Chen, Y.~Liu, J.~Hu, and Y.~Yuan, ``Dynamic depth-aware network for
  endoscopy super-resolution,'' \emph{JBHI}, vol.~26, no.~10, pp. 5189--5200,
  2022.

\bibitem{chen2024unsupervised}
\BIBentryALTinterwordspacing
K.~Chen, J.~Liu, R.~Wan, V.~H.-F. Lee, V.~Vardhanabhuti, H.~Yan, and H.~Li,
  ``Unsupervised domain adaptation for low-dose ct reconstruction via bayesian
  uncertainty alignment,'' 2024. [Online]. Available:
  \url{https://arxiv.org/abs/2302.13251}
\BIBentrySTDinterwordspacing

\bibitem{zhao2019channel}
X.~Zhao, Y.~Zhang, T.~Zhang, and X.~Zou, ``Channel splitting network for single
  mr image super-resolution,'' \emph{IEEE Trans. Imag. Process.}, vol.~28,
  no.~11, pp. 5649--5662, 2019.

\bibitem{li2019two}
Z.~Li, Q.~Liu, Y.~Li, Q.~Ge, Y.~Shang, D.~Song, Z.~Wang, and J.~Shi, ``A
  two-stage multi-loss super-resolution network for arterial spin labeling
  magnetic resonance imaging,'' in \emph{MICCAI}.\hskip 1em plus 0.5em minus
  0.4em\relax Springer, 2019, pp. 12--20.

\bibitem{almalioglu2020endol2h}
Y.~Almalioglu, K.~B. Ozyoruk, A.~Gokce, K.~Incetan, G.~I. Gokceler, M.~A.
  Simsek, K.~Ararat, R.~J. Chen, N.~J. Durr, F.~Mahmood \emph{et~al.},
  ``Endol2h: Deep super-resolution for capsule endoscopy,'' \emph{IEEE Trans.
  Med. Imag.}, vol.~39, no.~12, pp. 4297--4309, 2020.

\bibitem{upadhyay2019mixed}
U.~Upadhyay and S.~P. Awate, ``A mixed-supervision multilevel gan framework for
  image quality enhancement,'' in \emph{MICCAI}.\hskip 1em plus 0.5em minus
  0.4em\relax Springer, 2019, pp. 556--564.

\bibitem{chen2024mask}
W.~Chen, W.~Zhao, Z.~Chen, T.~Liu, L.~Liu, J.~Liu, and Y.~Yuan, ``Mask-aware
  transformer with structure invariant loss for ct translation,'' \emph{Medical
  Image Analysis}, vol.~96, p. 103205, 2024.

\bibitem{goodfellow2014generative}
I.~J. Goodfellow, J.~Pouget-Abadie, M.~Mirza, B.~Xu, D.~Warde-Farley, S.~Ozair,
  A.~C. Courville, and Y.~Bengio, ``Generative adversarial nets,'' in
  \emph{NIPS}, 2014.

\bibitem{yu2018crafting}
K.~Yu, C.~Dong, L.~Lin, and C.~C. Loy, ``Crafting a toolchain for image
  restoration by deep reinforcement learning,'' in \emph{CVPR}, 2018, pp.
  2443--2452.

\bibitem{zhang2018dynamically}
X.~Zhang, Y.~Lu, J.~Liu, and B.~Dong, ``Dynamically unfolding recurrent
  restorer: A moving endpoint control method for image restoration,'' in
  \emph{ICLR}, 2018.

\bibitem{furuta2019fully}
R.~Furuta, N.~Inoue, and T.~Yamasaki, ``Fully convolutional network with
  multi-step reinforcement learning for image processing,'' in \emph{AAAI},
  vol.~33, no.~01, 2019, pp. 3598--3605.

\bibitem{li2020mri}
W.~Li, X.~Feng, H.~An, X.~Y. Ng, and Y.-J. Zhang, ``Mri reconstruction with
  interpretable pixel-wise operations using reinforcement learning,'' in
  \emph{AAAI}, vol.~34, no.~01, 2020, pp. 792--799.

\bibitem{hui2021learning}
Z.~Hui, J.~Li, X.~Wang, and X.~Gao, ``Learning the non-differentiable
  optimization for blind super-resolution,'' in \emph{CVPR}, 2021, pp.
  2093--2102.

\bibitem{vassilo2020multi}
K.~Vassilo, C.~Heatwole, T.~Taha, and A.~Mehmood, ``Multi-step reinforcement
  learning for single image super-resolution,'' in \emph{CVPRW}, 2020, pp.
  512--513.

\bibitem{sutton2018reinforcement}
R.~S. Sutton and A.~G. Barto, \emph{Reinforcement learning: An
  introduction}.\hskip 1em plus 0.5em minus 0.4em\relax MIT press, 2018.

\bibitem{guo2021semantic}
X.~Guo, J.~Liu, and Y.~Yuan, ``Semantic-oriented labeled-to-unlabeled
  distribution translation for image segmentation,'' \emph{IEEE transactions on
  medical imaging}, vol.~41, no.~2, pp. 434--445, 2021.

\bibitem{veeling2018rotation}
B.~S. Veeling, J.~Linmans, J.~Winkens, T.~Cohen, and M.~Welling, ``Rotation
  equivariant cnns for digital pathology,'' in \emph{MICCAI}.\hskip 1em plus
  0.5em minus 0.4em\relax Springer, 2018, pp. 210--218.

\bibitem{he2016deep}
K.~He, X.~Zhang, S.~Ren, and J.~Sun, ``Deep residual learning for image
  recognition,'' in \emph{CVPR}, 2016, pp. 770--778.

\bibitem{dabov2007image}
K.~Dabov, A.~Foi, V.~Katkovnik, and K.~Egiazarian, ``Image denoising by sparse
  3-d transform-domain collaborative filtering,'' \emph{IEEE Trans. Imag.
  Process.}, vol.~16, no.~8, pp. 2080--2095, 2007.

\bibitem{gu2014weighted}
S.~Gu, L.~Zhang, W.~Zuo, and X.~Feng, ``Weighted nuclear norm minimization with
  application to image denoising,'' in \emph{CVPR}, 2014, pp. 2862--2869.

\end{thebibliography}
	
\end{document}